\documentclass[10pt,twocolumn,letterpaper]{article}

\usepackage[margin=1in]{geometry}
\usepackage{times}
\usepackage{graphicx}
\usepackage{booktabs}
\usepackage{amsmath,amssymb}
\usepackage[pagebackref=false,breaklinks=true,colorlinks=true,citecolor=blue]{hyperref}

\title{TellTale: Blending Multi-Instance LoRA Text Encoders and a\\
Zero-Shot LLM Judge for Ambivalence/Hesitancy Recognition in Videos}

\author{Abdel-Karim Al-Tamimi\\
Sheffield Hallam University, Sheffield, UK\\
Yarmouk University, Irbid, Jordan\\
{\tt\small a.al-tamimi@shu.ac.uk}
\and
Ali Rodan\\
The University of Jordan, Amman, Jordan\\
{\tt\small a.rodan@ju.edu.jo}}

\begin{document}
\maketitle

\begin{abstract}
We present TellTale, a text-only approach to ambivalence/hesitancy (A/H)
recognition in interview videos, evaluated on the BAH dataset as part of
the 3rd A/H Video Recognition Challenge (11th ABAW Workshop, ECCV 2026).
Although the dataset provides video, audio, facial crops, and transcripts,
TellTale relies on the transcript alone and combines three probability
streams. Two text encoders, multilingual-e5-large and mDeBERTa-v3-base,
are fine-tuned with parameter-efficient LoRA adapters under a
multiple-instance learning (MIL) objective, in which transcript chunks are
scored individually and pooled with a smooth maximum so that only the
video-level label is needed for supervision. The third stream requires no
training: a quantized 14B instruction LLM is prompted, zero-shot, to rate
each transcript for A/H. The three probabilities are combined by a
weighted average and a single decision threshold, both selected on
participant-grouped cross-validated predictions. On the organizer-scored
private test set of 152 videos from unseen participants, TellTale achieves
a Macro-F1 of 0.7364 and an average precision of 0.7940, compared with
0.2827 Macro-F1 for the official vision-based baseline. 
\end{abstract}

\section{Introduction}
\label{sec:intro}

Ambivalence and hesitancy (A/H) are mixed affective and cognitive states in
which a speaker holds conflicting attitudes toward what they are saying.
They may endorse a goal while hedging, correcting themselves, or struggling
to commit to a position. Detecting A/H automatically matters for
behavioural-change applications such as digital health coaching, where a
hesitant answer signals unresolved doubt. The BAH dataset~\cite{bah_dataset} was collected for this purpose. It contains
1{,}427 short videos of 300 participants answering interview questions,
each video labelled as showing A/H or not. The 3rd A/H Video Recognition
Challenge, held at the 11th ABAW Workshop~\cite{kollias2023abaw} and
building on a previous edition~\cite{abaw10ah}, poses the task as binary
video-level classification. Systems are evaluated by Macro-F1, the mean of
the two per-class F1 scores, on a private test set of 152 videos from 30
unseen participants. The average precision (AP) of the positive class is a
secondary metric. The official baseline, a zero-shot vision-only
Video-LLaVA model~\cite{lin2023videollava}, scores a Macro-F1 of 0.2827.

TellTale uses the transcript alone. During development we evaluated
visual, acoustic, and affect-based alternatives, both individually and
fused with text, and none of them improved over text-only models. The final
system therefore relies entirely on what the participant says. It blends
three transcript-based probability streams: two LoRA-fine-tuned text
encoders trained with a multiple-instance objective, and one zero-shot LLM
judge. On the private test set this blend reaches a Macro-F1 of 0.7364
(AP 0.7940).

Three properties make the method practical and easy to reproduce. First, it
is parameter-efficient: LoRA~\cite{hu2022lora} freezes each pretrained
encoder and trains only small low-rank adapter matrices, so the whole
system trains on a single consumer machine. Second, it is weakly supervised
at the right granularity. The label ``this video shows A/H'' does not say
where the hesitancy occurs, and multiple-instance
learning~\cite{dietterich1997mil,ilse2018attention} handles exactly this
situation by letting the model discover which transcript chunk carries the
evidence. Third, its final ingredient is training-free: an
instruction-tuned (Qwen3:14B) LLM~\cite{qwen3}, prompted once per video, supplies an
independent probability whose errors only partly overlap with those of the
fine-tuned encoders.

\section{Methodology}
\label{sec:method}

\subsection{Overview}

\begin{figure*}[t]
\centering
\includegraphics[width=0.88\textwidth]{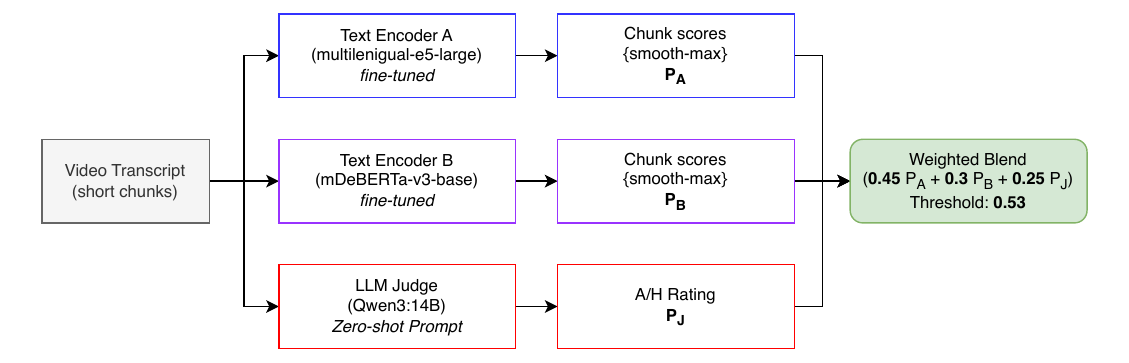}
\caption{The TellTale system. The interview transcript feeds three streams:
two fine-tuned text encoders whose per-chunk scores are pooled by a smooth
maximum into video probabilities $p_{\mathrm{A}}$ and $p_{\mathrm{B}}$, and
a zero-shot LLM judge whose 0--100 A/H rating becomes $p_{\mathrm{J}}$. A
weighted average of the three probabilities, followed by one threshold,
produces the decision. All weights and the threshold were chosen on
participant-grouped cross-validated predictions only.}
\label{fig:arch}
\end{figure*}

Figure~\ref{fig:arch} shows the system. Each transcript is split into short
chunks (the 3--5\,s segments provided with BAH, produced by a Whisper-style
recognizer~\cite{radford2023whisper}). Three streams each produce one
probability that the video shows A/H, and the final probability is their
weighted average:
\begin{equation}
p \;=\; 0.45\, p_{\mathrm{A}} + 0.30\, p_{\mathrm{B}} + 0.25\, p_{\mathrm{J}}.
\label{eq:blend}
\end{equation}
Here $p_{\mathrm{A}}$ and $p_{\mathrm{B}}$ come from the two fine-tuned
encoders (Section~\ref{sec:encoders}) and $p_{\mathrm{J}}$ from the LLM
judge (Section~\ref{sec:judge}). The video is predicted positive when $p$
reaches the decision threshold of 0.53. Neither the weights nor the
threshold are hand-picked. Section~\ref{sec:selection} explains how both
were derived from cross-validated predictions and why the resulting values
are reasonable.

\subsection{Evaluation protocol}
\label{sec:protocol}

Every selection decision in this paper uses 5-fold cross-validation with
participant grouping~\cite{stone1974cv,pedregosa2011sklearn} over the
1{,}427 labelled videos. All videos of a participant fall in the same fold,
so a model is never evaluated on a speaker it has seen during training.
Training on four folds and predicting the fifth, rotated five times, gives
each video exactly one prediction from a model that never saw it; these are
called out-of-fold (OOF) predictions. Architectures, blend weights, and
thresholds are chosen only on the pooled OOF predictions, never on the test
set. Within each training set, a further 10\% participant-grouped split is
held out for early stopping. The protocol proved reliable in practice,
private-test Macro-F1 landed within $\pm0.015$ of its OOF estimate for
every encoder-based configuration we evaluated.

\subsection{Streams A and B: MIL-fine-tuned text encoders}
\label{sec:encoders}

Both encoder streams share one idea: score every chunk, and let the most
A/H-like chunk decide the video. Each encoder carries a LoRA adapter and a
small classification head and produces one logit $z_t$ per chunk $t$. The
chunk logits of a video are pooled with a smooth maximum, the LogSumExp
function:
\begin{equation}
z_{\text{video}} = \frac{1}{r} \log \Bigl( \frac{1}{T} \sum_{t=1}^{T} e^{\,r z_t} \Bigr).
\label{eq:lse}
\end{equation}
Here $T$ is the number of chunks in the video and $r$ controls how closely
the function approximates a true maximum: at $r{=}1$ it is close to an
average, and as $r$ grows it converges to $\max_t z_t$ while remaining
differentiable. We use $r{=}50$, chosen because a sweep of pooling
functions on OOF predictions showed performance increasing steadily from
mean pooling toward max pooling and then saturating; $r{=}50$ captures the
max-like regime while keeping gradients defined for all chunks. We cap $T$
at 64 chunks, a bound that no video in the dataset reaches (the average is
about four chunks), so the cap exists only as a memory safeguard. The
pooled logit is trained against the video label with class-weighted binary
cross-entropy using the AdamW optimizer~\cite{loshchilov2019adamw}, so no
chunk-level labels are needed at this stage. With an identical encoder, averaging chunk
predictions reached an OOF Macro-F1 of 0.6484, whereas MIL fine-tuning
reached 0.7147.

Stream A uses multilingual-e5-large~\cite{wang2024e5}. It is first
pretrained at chunk level, where chunks of positive videos are labelled by
their time overlap with the dataset's annotated A/H segments, and is then
fine-tuned with the MIL objective of Eq.~\eqref{eq:lse}. Stream B uses
mDeBERTa-v3-base~\cite{he2023debertav3} and is trained with the MIL
objective directly. Being a smaller model trained from a different
pretraining recipe, it did not benefit from the extra chunk-level phase in
preliminary runs. Each stream is trained independently per fold and per
random seed, and its probability for a video is the average over all of its
checkpoints. Averaging across seeds matters because individual training
runs vary. The ensemble is what makes the reported numbers stable. On its
own, stream A reaches an OOF Macro-F1 of 0.7147 and stream B reaches
0.6666. Stream B earns its place through diversity rather than raw
strength since a differently pretrained encoder makes partly different errors,
which the blend can exploit.

Table~\ref{tab:hparams} lists all hyperparameters. The choices are
conventional and were kept deliberately simple. The LoRA rank of 16 (with
the standard scaling $\alpha = 2r$) was preferred over rank 8 after a
preliminary comparison showed a small improvement. We used ranks train under 1\% of
each encoder's parameters, which acts as regularization on a dataset of
this size. Dropout is slightly lower for stream B (0.05 versus 0.1)
because the smaller model is less prone to overfitting. The token limits
(256 and 128) comfortably cover the short 3--5\,s chunks. Learning rates
follow common practice for LoRA fine-tuning: a higher rate
($2{\times}10^{-4}$) for stream A's chunk-level phase, then a lower rate
($5{\times}10^{-5}$) for its MIL phase so the second phase adjusts rather
than overwrites what the first phase learned. Stream B trains its single
MIL phase at $2{\times}10^{-5}$, typical for DeBERTa-family models. Batch
sizes (16 chunks; 4 videos) are set by the memory of a single consumer
GPU. Exact epoch counts are not critical because early stopping on the
held-out grouped split terminates training when validation performance
stops improving. Stream A uses three seeds and stream B two, the primary
stream receives the larger ensemble, and the extra seed gave stream B no
further OOF improvement.

\begin{table}[t]
\centering
\caption{Hyperparameters of the two encoder streams. LoRA settings are
written as rank / scaling / dropout. Chunk-level pretraining applies to
stream A only. Both streams end with MIL training (Eq.~\eqref{eq:lse},
$r{=}50$, at most 64 chunks per video, class-weighted binary cross-entropy,
early stopping on a 10\% grouped split).}
\label{tab:hparams}
\resizebox{\columnwidth}{!}{%
\begin{tabular}{@{}lcc@{}}
\toprule
 & Stream A & Stream B \\
\midrule
Base encoder & multilingual-e5-large & mDeBERTa-v3-base \\
Parameters & $\sim$560M & $\sim$86M \\
Chunk embedding & mean-pooled & CLS token \\
Input prefix & \texttt{"query: "} & none \\
Max tokens per chunk & 256 & 128 \\
LoRA (rank/scale/drop) & 16 / 32 / 0.1 & 16 / 32 / 0.05 \\
\midrule
\multicolumn{3}{@{}l@{}}{\emph{Chunk-level pretraining (stream A only)}} \\
Epochs / LR / batch & 6 / $2{\times}10^{-4}$ / 16 & --- \\
\midrule
\multicolumn{3}{@{}l@{}}{\emph{MIL training (video-level, Eq.~\eqref{eq:lse})}} \\
Epochs / LR & 2 / $5{\times}10^{-5}$ & 6 / $2{\times}10^{-5}$ \\
Batch size (videos) & 4 & 4 \\
\midrule
Folds $\times$ seeds & $5 \times 3$ & $5 \times 2$ \\
Checkpoints averaged & 15 & 10 \\
OOF Macro-F1 (alone) & 0.7147 & 0.6666 \\
\bottomrule
\end{tabular}}
\end{table}

\subsection{Stream J: a zero-shot LLM judge}
\label{sec:judge}

The third stream involves no training. Qwen3:14B~\cite{qwen3}, quantized to
4 bits, is prompted once per video as a behavioural rater. The system
prompt defines A/H (conflicting attitudes, hedging, self-contradiction,
reluctance to commit), the user message contains the full transcript, and
the model replies in strict JSON with an integer A/H score from 0 to 100.
The video's probability is the score divided by 100. The prompt contains no
labelled examples, so this stream is independent of the fold structure and
cannot leak training labels into the evaluation. Scoring takes about five
seconds per video on one Apple-Silicon machine (MacBook Pro M3 with 64GB). On its own, this untrained
stream reaches an OOF Macro-F1 of 0.6970, close to the fully fine-tuned
stream A, and its errors overlap little enough with the encoders' to make
it a valuable third voice.

\subsection{Selecting the blend weights and threshold}
\label{sec:selection}

The weights in Eq.~\eqref{eq:blend} were found by an exhaustive grid search
over all weight triples $(w_{\mathrm{A}}, w_{\mathrm{B}}, w_{\mathrm{J}})$
that are non-negative, sum to one, and lie on a grid of step 0.05. Each
candidate triple was scored by the Macro-F1 of the blended OOF
probabilities at their best threshold, and the best triple, $(0.45, 0.30,
0.25)$, was kept. The step size is deliberately coarse, with 1{,}427
videos, as finer weight tuning starts fitting noise in the OOF predictions
rather than real differences between streams.

Moreover, the selected weights are interpretable. Stream A, the strongest individual
model, receives the largest share. Stream B receives a substantial 0.30
despite being clearly weaker alone, because the search values how different
its errors are from stream A's, not just its accuracy. The judge receives
0.25 for the same reason: it contributes an independent reading of the
transcript rather than a stronger one. The search did not concentrate all
weight on the best model, which is the expected outcome when the streams'
errors are partly independent, since a weighted average of diverse
predictors reduces variance without requiring any single predictor to
improve.

The decision threshold uses a conservative flat-region rule. Among all
thresholds between 0.30 and 0.70 (step 0.005) whose OOF Macro-F1 lies
within 0.005 of the maximum, we take the midpoint of the widest contiguous
interval, which is 0.53. The search range brackets the positive-class
prevalence, and the 0.005 tolerance matches the resolution at which OOF
differences are meaningful. On a 152-video test set, a threshold sitting on
a narrow peak of the OOF curve is fragile. The flat-region midpoint costs
nothing measurable on OOF and adds robustness.

Because weights fitted on the full OOF pool are mildly optimistic, we also
computed an honest estimate. The same search was refit on four folds and
applied to the held-out fifth, giving a Macro-F1 of 0.7213 against 0.7147
for stream A alone. The blend therefore improves over its best component
even under conservative evaluation. At test time, every checkpoint of
streams A and B scores all test videos and per-stream probabilities are
averaged, the judge rates each video once, and Eq.~\eqref{eq:blend} with
the 0.53 threshold gives the prediction.

\section{Results}
\label{sec:results}

Table~\ref{tab:results} reports the performance of the full system and of
its components on the organizer-scored private test set, alongside the
official baseline. The three-stream blend achieves a Macro-F1 of
\textbf{0.7364} (AP 0.7940). These component rows serve as an ablation study, 
demonstrating that omitting the judge or using only one encoder leads to a decline in Macro-F1 performance..

\begin{table}[t]
\centering
\caption{Private-test results (152 videos, 30 unseen participants) for the
full system and its components. OOF is the pooled 5-fold out-of-fold
estimate on the 1{,}427 labelled videos; blends show the conservative
(leave-one-fold-out) estimate.}
\label{tab:results}
\resizebox{\columnwidth}{!}{%
\begin{tabular}{@{}lccc@{}}
\toprule
System & OOF F1 & Private F1 & Private AP \\
\midrule
Stream A alone & 0.7147 & 0.7165 & 0.8119 \\
Streams A+J & 0.7179 & 0.7168 & 0.7989 \\
Streams A+B & 0.7141 & 0.7235 & \textbf{0.8137} \\
\textbf{Streams A+B+J (full)} & 0.7213 & \textbf{0.7364} & 0.7940 \\
\midrule
Official baseline~\cite{lin2023videollava} & --- & 0.2827 & --- \\
\bottomrule
\end{tabular}}
\end{table}

Two observations stand out. First, the evaluation protocol generalised effectively to 
unseen data, with each configuration's private Macro-F1 remaining within
$\pm0.015$ of its conservative OOF estimate, and the OOF ranking successfully identifying the 
strongest configuration in advance. The full blend was the
only configuration whose conservative estimate beat stream A alone (0.7213
against 0.7147), and it also performed best on the private set, finishing
0.0129 ahead of the next configuration. Second, differences of a few
thousandths, such as 0.7165 against 0.7168, are within evaluation noise at
$n{=}152$ and should not be over-interpreted. The larger gaps, such as the
blend's margin over its components, are the meaningful ones.

\section{Discussion}
\label{sec:discussion}

In this interview setting, A/H is
expressed mainly in what is said. The clearest signs are hedging,
self-correction, qualifying language, and an inability to commit. Visual
and acoustic features did not add measurable signal in our development
experiments. We do not claim that visual A/H cues are absent. BAH annotates
gaze and facial cues, but exploiting them likely requires temporal video
models and more data than this task provides.

The video label does not say where the hesitancy occurs. Averaging chunk predictions dilutes one
hesitant utterance with otherwise unremarkable speech. The smooth-maximum
objective avoids this by letting the single most A/H-like chunk determine
the video score, and the encoder itself learns under that assumption. This
change of objective alone was worth $+0.066$ OOF Macro-F1 over averaging
with an identical encoder.

Each stream reads the same transcript through
a different lens. Two differently pretrained encoders and one
instruction-tuned LLM make partly different mistakes, and those mistakes
cancel when their outputs are averaged. This is exactly what the weight
search rewarded (Section~\ref{sec:selection}). Under a
conservative leave-one-fold-out evaluation, blend was the only configuration
that beat the best single stream, and the same ordering held on the private
test set.

\section{Conclusions}
\label{sec:conclusions}

TellTale combines two LoRA-fine-tuned text encoders, trained with a
video-level multiple-instance objective, with a zero-shot LLM judge. On the
private test set of the 3rd A/H Video Recognition Challenge it achieves a
Macro-F1 of 0.7364 and an AP of 0.7940 using transcripts alone. The
official vision-based baseline scores a Macro-F1 of 0.2827 on the same
data. The proposed approach is designed to be simple to replicate. It needs two
encoders, as detailed in (Table~\ref{tab:hparams}), one pooling equation, one prompt, one
blend, and one threshold. The evaluation protocol behind these choices
predicted private-test performance within $\pm0.015$ Macro-F1 for every
configuration evaluated. Future work includes exploiting the dataset's
visual cue annotations with temporal video models and testing whether the
blend's advantage persists at larger dataset scales.

{\small
\bibliographystyle{ieeetr}
\bibliography{refs}
}

\end{document}